RESEARCH **Open Access**

# IGRF-RFE: a hybrid feature selection method for MLP-based network intrusion detection on UNSW-NB15 dataset

Yuhua Yin[1*], Julian Jang-Jaccard[1], Wen Xu[1], Amardeep Singh[1], Jinting Zhu[1], Fariza Sabrina[2] and Jin Kwak[3]

*Correspondence:
yuhua.yin.1@uni.massey.ac.nz

[1] Comp Sci/Info Tech, Cybersecurity Lab, Massey University, Auckland, New Zealand
[2] School of Engineering and Technology, Central Queensland University, Sydney, Australia
[3] Department of Cyber Security, Ajou University, Suwon, Republic of Korea

**Abstract**

The effectiveness of machine learning models can be significantly averse to redundant and irrelevant features present in the large dataset which can cause drastic performance degradation. This paper proposes IGRF-RFE: a hybrid feature selection method tasked for multi-class network anomalies using a multilayer perceptron (MLP) network. IGRF-RFE exploits the qualities of both a filter method for its speed and a wrapper method for its relevance search. In the first phase of our approach, we use a combination of two filter methods, information gain (IG) and random forest (RF) respectively, to reduce the feature subset search space. By combining these two filter methods, the influence of less important features but with the high-frequency values selected by IG is more effectively managed by RF resulting in more relevant features to be included in the feature subset search space. In the second phase of our approach, we use a machine learning-based wrapper method that provides a recursive feature elimination (RFE) to further reduce feature dimensions while taking into account the relevance of similar features. Our experimental results obtained based on the UNSW-NB15 dataset confirmed that our proposed method can improve the accuracy of anomaly detection as it can select more relevant features while reducing the feature space. The results show that the feature is reduced from 42 to 23 while the multi-classification accuracy of MLP is improved from 82.25% to 84.24%.

## Introduction

The Internet has changed the way people communicate, work, build businesses, and live our daily life dramatically. However, with the increasing number of network connections and network services, network attacks have become a major challenge for human society. According to Norton's annual security report published in 2021, a network attack occurs every 39 s globally [1]. In terms of attack forms, network attacks can be categorized into active attacks and passive attacks [2]. Active attacks can have great impacts on system usability, and the most typical example is a denial of service attack. Passive attacks aim to capture important information in computer systems.

To mitigate the risk of different types of attacks, intrusion detection systems have been developed to detect malicious behaviors in the network [3, 4]. An early intrusion detection system was proposed by Denning in 1987, who described a model based on audit records and statistical methods to identify system anomalies [5]. Modern intrusion detection systems can be mainly divided into three categories, which are signature-based, anomaly-based, and hybrid [6]. Signature-based IDS matches different types of attacks against a pre-specified database of signatures. One of its disadvantages is that it cannot effectively detect unknown attacks because of Zero-day attacks and outdated databases. Anomaly-based IDS detects attacks by learning normal and anomalous network behaviors and has better detection capabilities for unknown attacks. However, due to the problems of redundant features and class imbalance in intrusion detection datasets, anomaly-based IDS has been shown to lead to false positives. Hybrid IDS combines signature-based and anomaly-based approaches [7].

The network intrusion detection tasks have become more complex in recent years as new network attacks continue to emerge and network data traffic increases. Consequently, machine learning has been widely used in intrusion detection systems because of its ability to learn and identify patterns from complex data through statistical methods and advanced algorithms [8]. Intrusion detection methods based on machine learning can be divided into two categories: supervised learning and unsupervised learning. In supervised learning, machine learning methods such as decision trees and random forests classify network behavior by learning from the labeled data [9]. Unsupervised intrusion detection methods such as K-means and hidden Markov models focus on the clustering problem [10] to group network behaviors [11].

Deep learning is a major branch of machine learning that is based on neural networks with at least two hidden layers. Deep learning is better suited at automatically learning and extracting features from large data sets, and has shown promising performance [8, 12–15]. In spite of these advantages, feature engineering still plays an important role in deep learning models when faced with high dimensional structured data [16]. High-dimension, redundant and irrelevant features may make the model overfitting during the learning process and result in a high false positive rate in the real network environment [17]. There has been a wide range of research applying different feature selection methods to assist the Intrusion Detection System (IDS) to improve performance and reduce the false positive rate.

A single feature selection method is based on the assumption of importance indicators to eliminate unimportant features. For example, information gain uses the information entropy between features and labels as feature importance indicators, while random forest judges the importance of features based on multiple decision trees. To avoid biased feature importance metrics, using hybrid feature selection methods can combine different metrics to prevent removing important features. Hsu et al. also pointed out that hybrid feature selection approaches would achieve more stable performance than a single feature selection method [18]. The purpose of this paper is to propose a hybrid feature selection method to help improve the multi-classification performance of intrusion detection systems on UNSW-NB15 dataset. We proposed a hybrid feature selection method named IGRF-RFE which combines both filter and wrapper methods that can reduce feature subset search space and eliminates redundant features.

The contributions of our work are as follows:

- We proposed an MLP-based intrusion detection system using a novel hybrid feature selection method called IGRF-RFE. IGRF-RFE is composed of both filter feature selection and wrapper feature selection methods. In the first step, it applies the ensemble feature selection method based on Information Gain and Random Forest Importance to reduce the dimension of features. This step can reduce the feature subset to a reasonable range while referring to two different feature importance metrics. Then, recursive feature elimination(RFE) as a wrapper method is applied to the reduced feature subset to remove features that negatively impact the actual model performance. An MLP classifier with two hidden layers is used in the RFE and the final classifier.
- Since there are many duplicate data in UNSW-NB15, feature selection can not be directly applied as this can cause overfitting. We analyzed the intrusion detection dataset and removed duplicated data before ranking the features to avoid ranking bias typically associated with selecting features that can cause overfitting.
- During data pre-processing, we removed the minority classes for UNSW-NB15. In addition, we employed a resampling technique to ensure a balance between normal and abnormal classes. This can avoid another type of overfitting problem typically associated with the classes with the limited number of samples available for training.
- The experimental results obtained based on the UNSW-NB15 dataset showed that our proposed model can reduce feature dimension from 42 to 23 while achieving a detection accuracy of 84.24% compared to 82.25% before feature selection.

We organized the rest of the paper as follows. In section "Related works", we discussed related works on feature selection methods for intrusion detection systems. In section "Proposed method", we introduced our MLP-based intrusion detection system as well as IGRF-RFE feature selection methods. In section "Experiments and results", we presented our experiment details and results. The conclusion and future works were present in section "Conclusion and future work".

## Related works

In machine learning, feature selection is an important measure that can help eliminate low-value features, avoid overfitting, reduce detection time and improve model accuracy. Defined by methodologies, feature selection methods can be divided into three categories: filter methods, embedded methods, and wrapper methods [19]. Filter methods can rank features based on some metrics such as statistical measures, information distance, and correlations for example to select the best-ranked features [20]. As filter methods are model-independent, feature importance is consistent and does not require recalculation. Embedded methods obtain feature importance scores from tree-based machine learning algorithms such as random forest, C4.5, and Xgboost. After ranking the features by importance, similar to filter methods, forward feature search or backward feature elimination can be applied to select feature subsets [21]. Wrapper methods evaluate the quality of feature subsets based on their actual performance on machine learning models [22]. Wrapper methods are not model-independent and thus can be based on any models. Wrapper methods perform actual training on the model for each evaluation of a feature subset, as a result, they are more time and computational consuming than filter methods. To reduce selection time, random search algorithms or other methods are typically used together with the wrapper method.

Zhou et al. proposed a feature selection method CFS-BA for intrusion detection systems, which was based on correlation feature selection and bat algorithm [23]. The purpose of this method was to find the least relevant feature subset through an optimized random search algorithm. In this study, an ensemble voting classifier based on random forest, C4.5, and Forest PA were used, and experiments were performed on three datasets NSL-KDD, AWID, and CIC-IDS2017. The results showed that CFS-BA could reduce the number of features of these three datasets to 10, 8, and 13, and improved accuracy by 4.5%, 1.3%, and 2.2% in binary classification respectively.

The researchers in [24] proposed a filter feature selection method using the Gini index for intrusion detection systems and used the GBDT model as the classifier. In this study, the PSO algorithm was also used to find the optimal hyper-parameters for GBDT. To verify the effectiveness of this model, the author applied the model to the NSL-KDD dataset, and the Gini index method reduced the number of features from 41 to 18. The optimized GBDT classifier could achieve a performance of 86% in accuracy and 3.83% in false positive rate.

Kasongo et al. used Xgboost as an ensemble feature selection method for intrusion detection systems in their research and made a performance analysis on the UNSW-NB15 dataset using machine learning models [25]. According to the feature importance ranked by Xgboost, the researchers selected the 19 most important features from the 42 features. The results showed that in the binary classification based on decision Trees, Xgboost feature selection improved the accuracy by 1.9% compared with the baseline performance using all features.

Eunice et al. proposed an intrusion detection system using random forests and deep neural networks (DNN) [26]. Their experiments used random forests to select different numbers of features and then used them in different layers of DNN. The experimental results showed that the best binary classification accuracy is 82.1% when 20 features were selected, and the DNN layer was 4. However, their experiments did not consider multi-classification performance under their proposed model.

In Prasad et al.'s work, a multi-level correlation-based feature selection was proposed in the intrusion detection systems on the UNSW-NB15 dataset [27]. In the two-level feature selection approach, Pearson correlation was used to evaluate feature-to-feature and feature-to-label correlations. If a pair of features' correlations were larger than 0.9, the redundant feature with a more significant mean absolute correlation was removed. In addition, feature-to-label correlation metrics were used for importance filtering. The experiment finally selected 15 features for a decision tree model and achieved a multi-classification accuracy of 95.2%. In their work, instead of the pre-prepared 10% training and test sets, they used the full dataset.

In the research by Alazzam et al. [28], a feature selection method based on Pigeon Inspired Optimizer (PIO), inspired by the behavior of pigeon groups, was proposed. In the study, the author proposed an improved PIO algorithm based on cosine similarity named Cosine PIO and compared it with Sigmoid PIO. The NSL-KDD, KDDCup99, and UNSW-NB15 datasets were used in the experiments. In binary classification, Cosine PIO performed better than Sigmoid PIO in all three datasets. It selected 5 features in NSL-KDD, 7 features in KDDCup99, 5 features in UNSW-NB15, and achieved an accuracy of 88.3% in NSL-KDD, 96% in KDDCup99, and 91.7% in UNSW-NB15.

Zhang et al. used Information Gain and ReliefF feature selection methods in a random forest-based intrusion detection system [29]. They conducted three sets of experiments on the NSL-KDD dataset. The researchers first examined the performance using Information Gain and ReliefF alone and then compared it with their combined method IG-ReliefF. The IG-ReliefF method could first use IG to reduce the dimension of features and then used the ReliefF method to rank the importance, which could effectively reduce the time and computational requirements required for feature selection. The experimental results showed that ReliefF could achieve higher accuracy than the individual IG and ReliefF methods.

Megantara et al. implemented a hybrid feature selection method based on Gini importance and recursive feature elimination (RFE) on the NSL-KDD dataset [30]. Gini importance was used as a filter method to rank feature importance, and RFE was used to further optimize the number of features through a decision tree-based wrapper method. Using the decision tree as a classifier, DOS, Probe, R2L, and U2R classes achieved 88.98%, 91.18%, 81.29%, and 99.42% accuracy in performance separately.

In [31], Ustebay et al. used a random forest-based recursive feature elimination algorithm in CICIDS2017 which contained 80 features. The experiment evaluated the results of selecting 1 to 80 features using recursive feature elimination. The 4 most important features Source Port, Flow Packet/s, Flow IAT Mean, and Flow IAT Std were identified, and the MLP-based IDS achieved 89% accuracy in performance. Because only a small part of the dataset was used for training in the experiment, the performance was not high, but the small data could reflect the generalizability of the model and feature in the real network environment.

Zong et al. implemented IG-TS, a random forest-based two-stage IDS on the UNSW-NB15 dataset [32]. The Information Gain feature selection method was used to reduce features and SMOTE was used to oversample the minority class. The first stage of the model focused on minority class classification, and the second stage focused on majority class classification. Combining the results of the two stages, IG-TS could achieve an accuracy of 85.78%.

Kumar et al. proposed an ensemble intrusion detection system based on multiple tree models (C5, CHAID, CART, QUEST) [33]. In the study, the authors used the UNSW-NB15 dataset for training and generated a real-time dataset to evaluate the performance of the model against unknown attacks. Through Information gain feature selection, the author reduced the number of features to 13. Then they used the reduced features to detect five majority classes which were DoS, Probe, Normal, Generic, and Exploit. The model achieved 83.4% accuracy in performance, evaluated in a real network environment.

These existing works exhibit a number of limitations. When decision trees or random forests are used alone. These techniques usually assign the same importance to correlated features but should have been a fraction, affecting the model interpretability. A combination of filter methods can address the disadvantage of a single method thus producing better outcomes. Also, the majority of the filter methods used in the existing state-of-the-art are generally univariate which ranks each feature independently of the rest. As a result, they tend to ignore any interaction that occurs between features thus often redundant variables are not eliminated. To address this issue, a wrapper method can be used to supplement the limitations of the univariate nature of the filter method. A wrapper method provides learning-based feature selection after evaluating the pros and cons of the features. When training (for feature selection), a wrapper method has the capability to take into account the relevance of features across the same feature subset space. This capability provides a more enhanced feature selection when the relative relevance across features should be accounted for.

## Proposed method

In this section, we introduced the overview of our proposed model—shown in Fig. 1. The UNSW-NB15 dataset contains 39 numerical features and 3 categorical features and provides a training set and a test set. Since it cannot be used in the MLP model directly, data pre-processing is applied to encode the dataset. During data preprocessing, we performed techniques including data cleaning, minority removal, oversampling, encoding, and normalization of the dataset. After data preprocessing, we divided the dataset into a training set, a validation set, and a test set. The training set and validation set are used in the feature selection and training process while the test set is used to verify the final performance of the model. Our proposed method has two steps. First, we applied an ensemble feature selection method based on information gain and random forest importance to filter important features. Then we performed recursive feature elimination on the reduced features to further optimize the feature subset. After feature selection, we used the obtained optimal feature subset to train the MLP model. The final performance on the test set provided the effectiveness of our proposed model.

### Ensemble feature selection with information gain and Random Forest Importance

#### *Information gain*

Information gain (IG) is a univariate filter feature selection method based on information entropy [34]. Entropy is a concept in information theory proposed by Shannon [35] and is often used to measure the uncertainty of a variant. When dealing with high-dimensional datasets, there may exist features that are highly skewed or contain little information, which affects machine learning performance. In classification problems, IG feature selection takes the amount of information as the importance metric by calculating the information entropy of each feature. As defined in Eq. 1, the information gain of a feature is equal to the entropy of the class label minus the conditional entropy of the class label under the feature. The formulas for class feature entropy and conditional class entropy are defined in Eqs. 2 and 3. After calculating the information gain for each feature, they can be ranked and selected according to the importance metric.

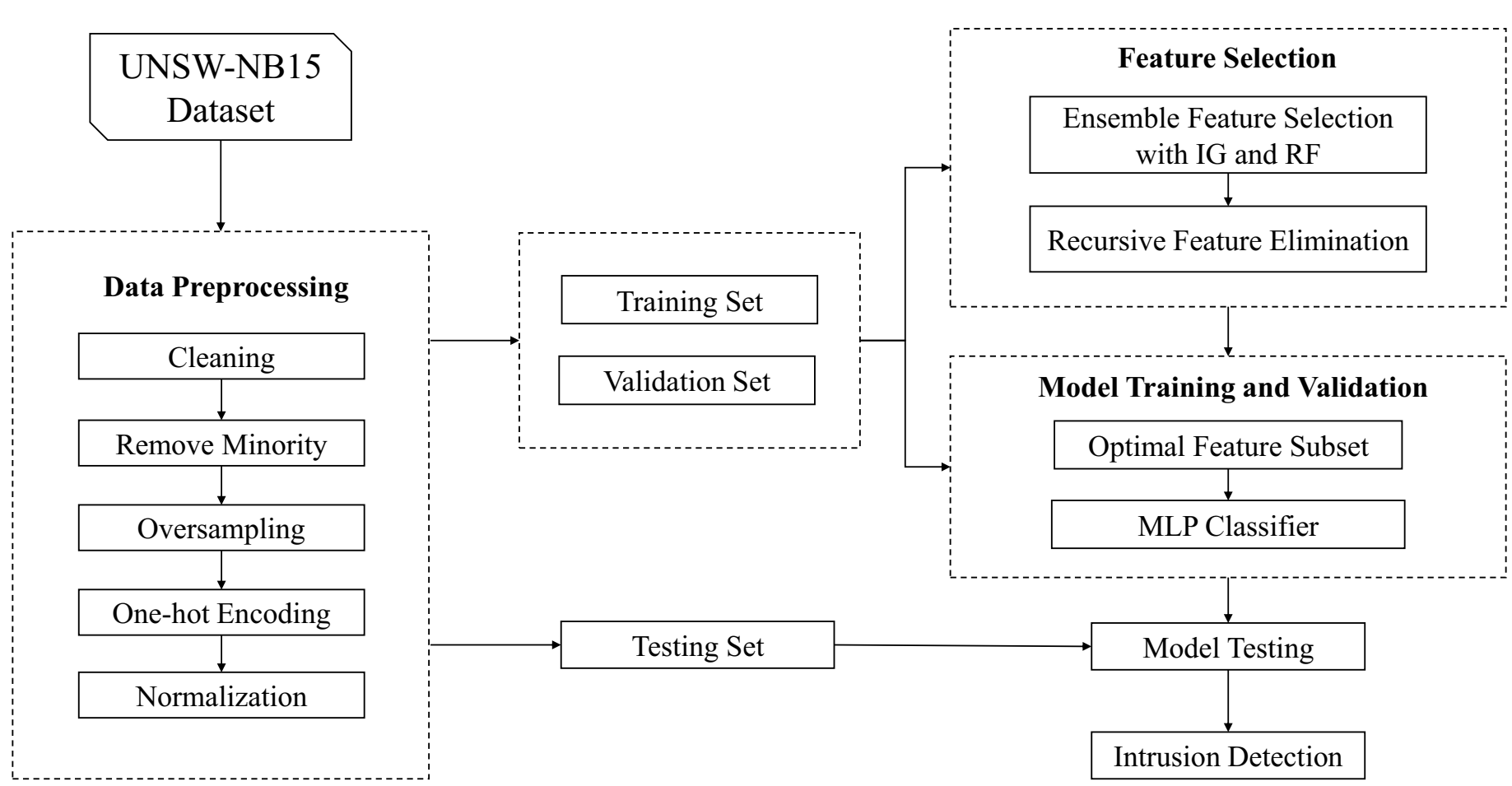

**Fig. 1** Our proposed model

$$IG(Y,X) = H(Y) - H(Y|X) \tag{1}$$

In Eq. 7, $Y$ represents the class vector and $X$ represents a feature vector.

$$H(Y) = -\sum_{i=1}^{n} p\left(y_i\right) log_2 p\left(y_i\right) \tag{2}$$

In Eq. 8, $n$ is the number of classes in vector $Y$ and $p(y_\mathrm{i}$ represents the probability of class $y_\mathrm{i}$ in class vector $Y$.

$$H(Y|X) = -\sum_{i=1}^{m} p(x_i) H(Y|X = x_i) \tag{3}$$

In Eq. 9, $m$ is the number of values contained in the feature vector $X$ and $p(x_\mathrm{i})$ represents the probability of value $x_\mathrm{i}$ in the feature vector $X$.

#### *Random forest feature importance*

Random forest is a machine learning method based on multiple decision trees, which is often used for many regression and classification tasks [36]. Different from decision trees, random forests add randomness to multiple decision trees to avoid overfitting and have better generalizability. When a random forest is used as a classifier (see Fig. 2), it first determines how many trees to build, and then uses the bootstrap sampling technique to randomly select a subset of the data for each decision tree. Another part of the randomness comes from the features used by each decision tree, and random feature subsets also lead to better generalizability and robustness. After training, the random forest classifier can generate the prediction with a higher probability by a voting method based on the prediction of each tree.

The random forest can also be used as an embedded feature selection method. The model can produce the importance score for each feature, which can be used to select the most important features and remove features that are not important for performance. The feature importance of random forest mainly depends on the node impurity property in decision trees. When generating a node in a decision tree, a feature's position and priority are determined based on the Gini index or entropy in a node. The lower Gini index or entropy represents less impurity and higher importance. The feature importance of random forests calculates the impurity of each feature in each tree and can get an average importance score (see Algorithm 1).

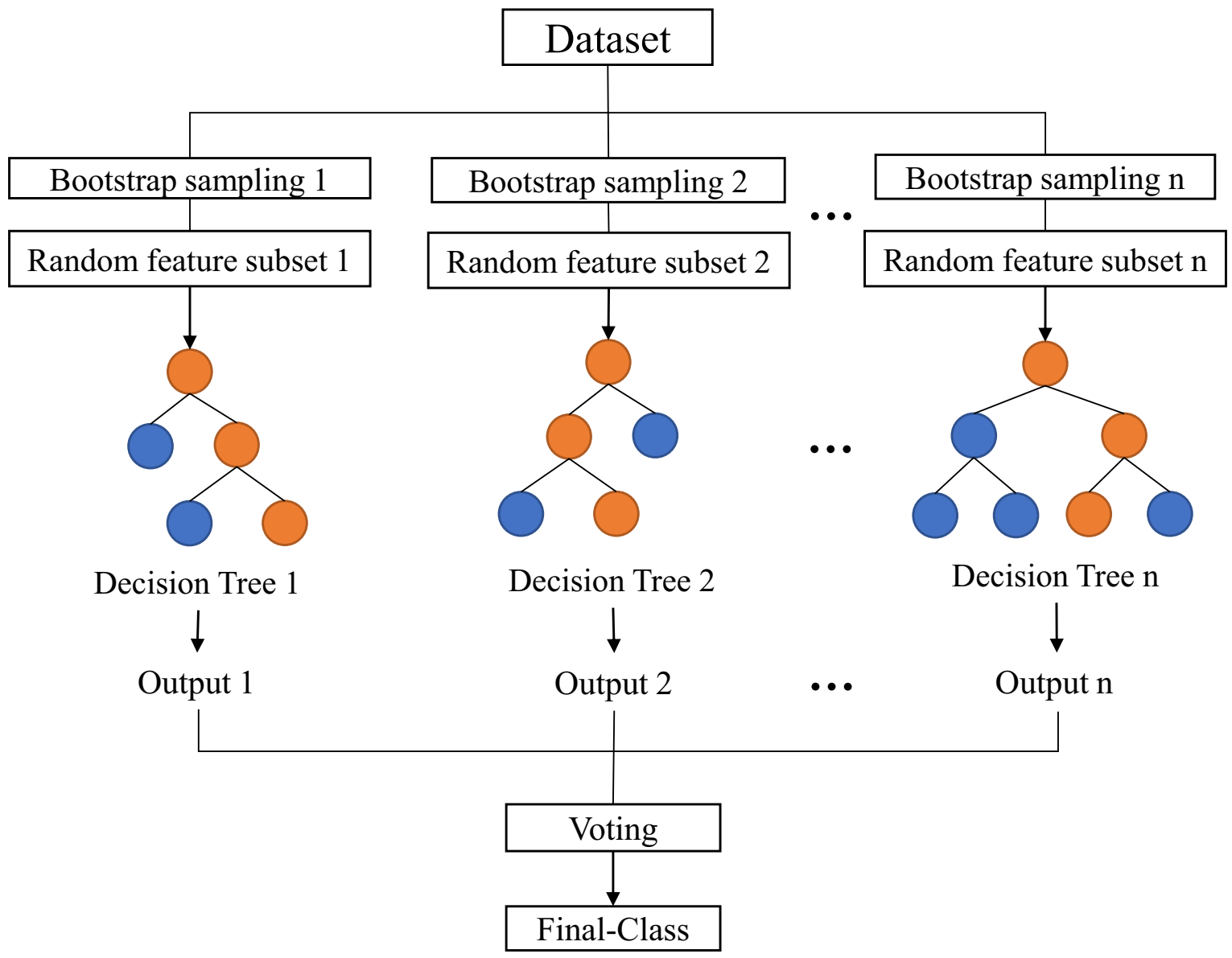

**Fig. 2** Random forest classifier

**Algorithm 1:** Random Forest Feature Importance

```
T: trees in random forest {t_1, t_2, ..., t_m}
F: features in dataset {f_1, f_2, ..., f_n}
for i from 1 to n do
    for tree t ∈ T do
        N: nodes using feature f_i in tree t {n_1, n_2, ..., n_p}
        for node n ∈ N do
            compute impurity decrease at n as a score s.
            weight the score s by number of samples.
            add up the score s to score S.
        end
    end
                                        /* get importance for feature f_i */
    f_i_importance = average score S over all trees t using feature f_i.
end
```

### *Ensemble feature selection*

Ensemble feature selection with information gain (IG) and random forest (RF) importance is the first step of our feature selection method. In this step, the ensemble feature selection method is only applied to 39 numerical features, and 3 categorical features are preserved to avoid loss of important information. As seen in Fig. 3, it first pre-processes the input training set to remove duplicate data. Duplicate data may reduce the generalizability of the result because selected features may overfit classes or instances with more repetitions. Subsequently, it calculates the importance of each feature using information gain and random forest respectively. The importance scores are normalized to a value between 0 and 1. By ranking and visualizing the importance scores, thresholds are selected to differentiate obviously unimportant features and other features [37]. A feature is retained if its importance is larger than the threshold, while it is removed if its importance is lower than the threshold. We assume that significant features could exist in both reduced feature subsets selected based on IG and RF metrics, so their union set is used for further feature optimization.

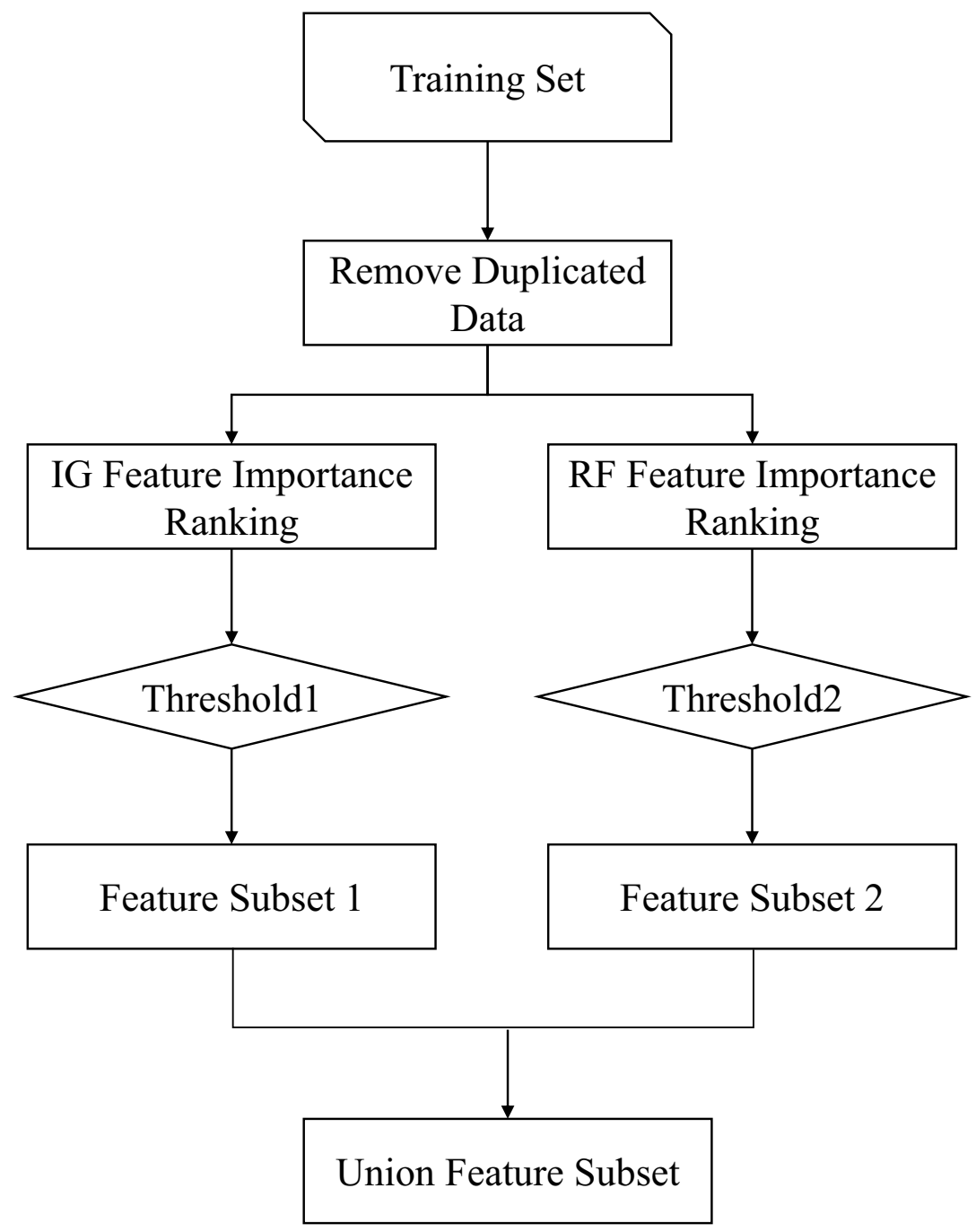

**Fig. 3** Workflow of ensemble feature selection with IG and RF

### Recursive feature elimination

Recursive feature elimination(RFE) is the second step in our feature selection method. RFE is a wrapper feature selection method, which can evaluate the importance of features iteratively based on machine learning performance by recursively eliminating each feature [38]. RFE removes the least important features in each iteration until the best performance is obtained or a specified number of features is reached. In our RFE algorithm (see Algorithm 2), the input training set and validation set contain only the reduced numeric features from the first stage and all categorical features. Other inputs to the algorithm include a positive integer patient $p$ and a list *init_features* containing selected features in the first stage. Patient $p$ is introduced to stop RFE in time if better performance cannot be obtained in several iterations, while *init_features* can be used to reduce the search space of RFE without starting from all features. Before recursive feature elimination, variables must be initialized. $f_len$ represents the number of initial features, which determines the iterations of RFE in the worst case. *best_performance* is used to record the best performance during RFE, *keep_features* stores the features selected after each iteration of the RFE, *selected_features* stores the feature subset for best performance, and *rm_list* stores removed numeric features during RFE. In the process of recursive feature elimination, an empty dictionary performance_dict will be initialized at the start of each iteration to store the validation performance with MLP after eliminating each feature. In the evaluate elimination function, the score is calculated by averaging the accuracy of 10 different experiments each of which is set with a different random seed. Subsequently, patient $p$ determines whether to continue RFE. If patient $p$ is larger than 0, one iteration of RFE is performed, and the local best performance of the iteration is obtained. After comparing the local best performance and global best performance, global best performance and selected features are updated.

**Algorithm 2:** Pseudocode of RFE with MLP

```
Input:
Training set T, Validation set V
patient p                                   /* monitor training performance */
    init_features = [f_1, f_2, f_3, ..., f_n]   /* a list of original features */
Output:
selected_features                           /* a list of selected features */
begin
                                            /* Recursive Feature Elimination */
    for init_features do
        if p > 0 then
            for keep_features do
                        /* take a feature from the list of features to be
                evaluated */
                temp_rm.append(keep_features[j])
                        /* evaluate the feature performance with MLP */
                score=evaluate_elimination(T,V,temp_rm)
                                /* store the score of each elimination */
                performance_dict[keep_features[j]]=score
            end
            /* get the feature name of best performance during elimination
            */
            max_key=get_max_key(performance_dict)
            rm_list.append(max_key)
            keep_features.remove(max_key)
            /* check if the performance of elimination is better than best
            performance */
            if performance_dict[max_key] > best_performance then
                best_performance=performance_dict[max_key]
                selected_features=keep_features
                increase patient counter
            else
                reduce patient counter
            end
        else
            return selected_features
        end
    end
end
```

## MLP classifier

### *Multilayer perceptron (MLP)*

MLP is a feed-forward artificial neural network with multiple hidden layers [39] (see Fig. 4). For classification problems, the amount of neurons in the output layer of MLP is equal to the number of classes to be classified while the number of neurons in the input layer is associated with the number of features. The layers between the input and output layers are often fully connected layers and are trained by backpropagation. When performing forward propagation, the network calculates the output of each layer based on an activation function from the previous layer as well as corresponding weight and bias values, as shown in Eq. 4.

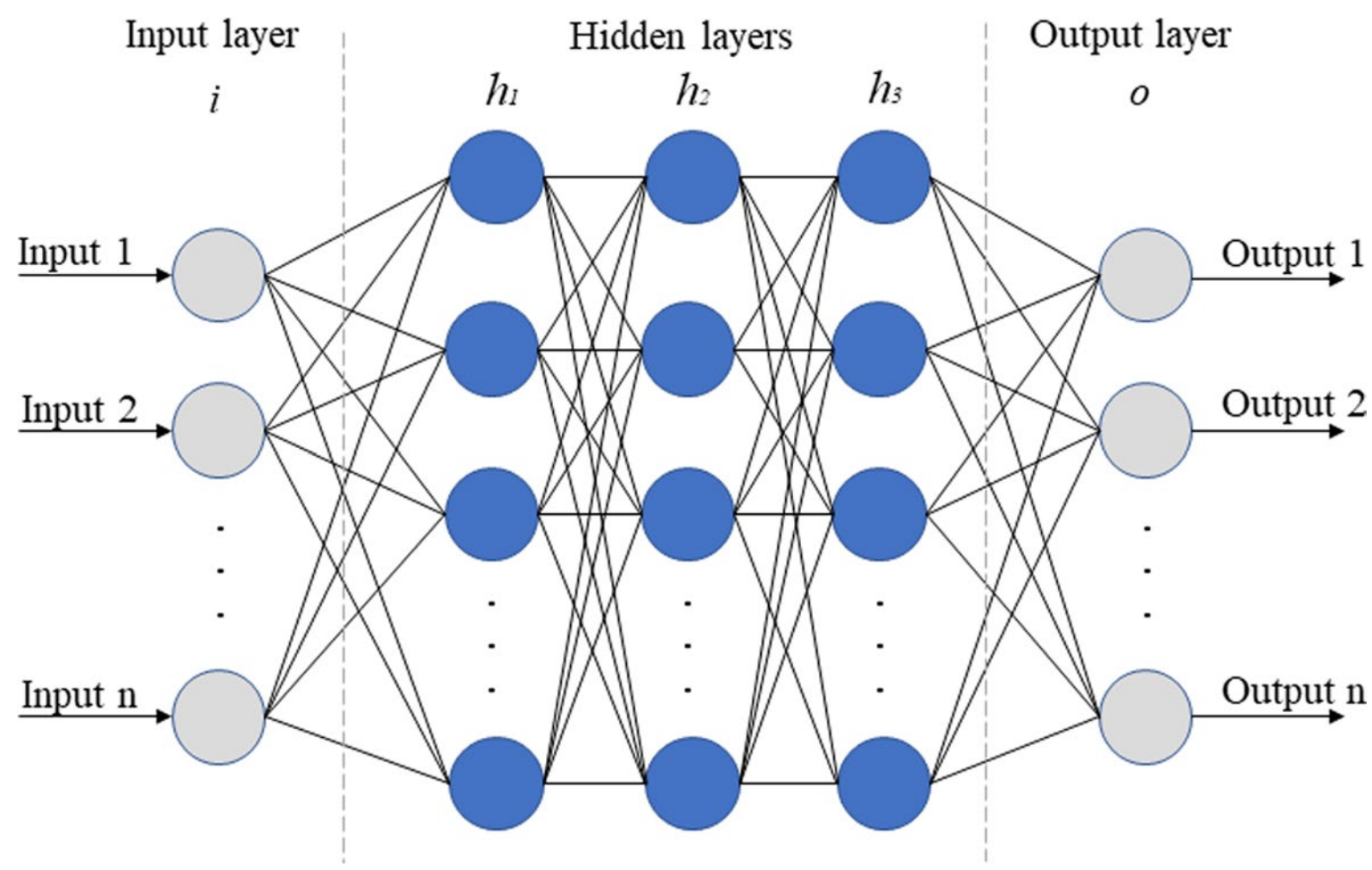

**Fig. 4** Basic MLP model

$$Z^{[l]} = W^{[l]}A^{[l-1]} + b^{[l]} \tag{4}$$

where $Z^{[l]}$ represents the output matrix, $W^{[l]}$ is weight matrix and $b^{[l]}$ represents bias vector.

Since the output of an MLP could be any value, an activation function is used to normalize the output. The activation function can transform the output of each layer to a certain range as shown in Eq. 5.

$$A^{[l]} = g\left(Z^{[l]}\right) \tag{5}$$

where $A^{[l]}$ represents the activated output matrix.

In our proposed method, we used Relu as the activation function for the hidden layer and Softmax as the activation function for the final output layer. Relu, as defined in Eq. 6, is an activation function that only transforms values less than zero to 0. The Softmax activation function, as defined in Eq. 7, is usually used for multi-classification, which can improve the defects of the sigmoid function for multi-classification, and ensure that the probability sum of the output layer is equal to 1. It can help determine the most probable prediction.

$$a_{\text{relu}} = max(0, z) \tag{6}$$

$$a_{\text{soft}} = \frac{e^{z_i}}{\sum_{j=1}^{J} e^{z_j}} \tag{7}$$

where J is the class number, $z_{\text{i}}$ represents the ith output value

The loss function as defined in Eq. 8 is used to calculate the error between the predicted value and the actual value, and then use back-propagation to adjust the weights w and bias b.

$$L(y,\hat{y}) = \frac{1}{m}\sum_{i=1}^{m}(y_i - \hat{y}_i)^2 \tag{8}$$

where m is the number of samples, $\hat{y}$ is the predicted value, and y is the exact value.

#### *Batch normalization*

For deep learning models, it is important to avoid overfitting. In a deep neural network, if the layers are too deep, it is possible to have gradient vanishing or gradient explosion problems, which may affect the performance of the model and may cause overfitting. Batch normalization, as defined in Eq. 9, is a method proposed by Loffe and Szegedy [40] to solve the gradient explosion or gradient vanishing. After each hidden layer, batch normalization normalizes the correspondent output values to avoid values that are too large or too small. It first takes the difference between each output and the vector's mean value and then divides it by a standard deviation. In this study, batch normalization is added after each hidden layer of our MLP model to avoid overfitting.

$$X_\mathrm{i} = \frac{X_\mathrm{i} - Mean_\mathrm{i}}{StdDev_\mathrm{i}} \tag{9}$$

where $X_i$ is the ith hidden layer's output matrix, $Mean_i$ is the mean value of $X_i$, and $StdDev_i$ is the standard deviation of $X_i$.

#### *Classification*

In this study, we implemented the MLP as a classifier with two hidden layers using the Relu activation function, each of which contains 128 neurons (see Fig. 5). After each hidden layer, batch normalization is added as a means of regularization. The selected features and pre-processed data are fed into the neural network through the input layer, the model is trained through forward and backward propagation and the output layer produces the probability of each class using the Softmax activation function. In the prediction stage, after producing a class probability vector, the argmax function, as defined in Eq. 10, finds the largest number among them and returns its index.

$$result = argmax(probability_vector) \tag{10}$$

Our model was trained with Adam's optimization algorithm, which adaptively adjusts the learning rate based on recent gradients for the weight. Also, our model used the learning rate = 0.0003, the batch size = 64, and the epochs = 300. To avoid overfitting, we apply the early-stopping technique, which can stop training in time when overfitting is observed, and restore the best model parameters. We set the parameter of early-stopping = 30. If the loss of the validation set does not decrease for more than 30 consecutive epochs, it is determined that the model has been overfitted thus the training stops and any changes are rolled back.

### Computational complexity

By analyzing the computational complexity of the single feature selection algorithm and our proposed hybrid feature selection, it can be found that the worst-case computational complexity of the two feature selection algorithms used in the first step, IG and RF, is $O(n)$ (see Eqs. 11 and 12). The worst-case computational complexity of our proposed IGRF-RFE is the same as that of RFE, which is $O(n^2)$ (see Eqs. 13 and 14).

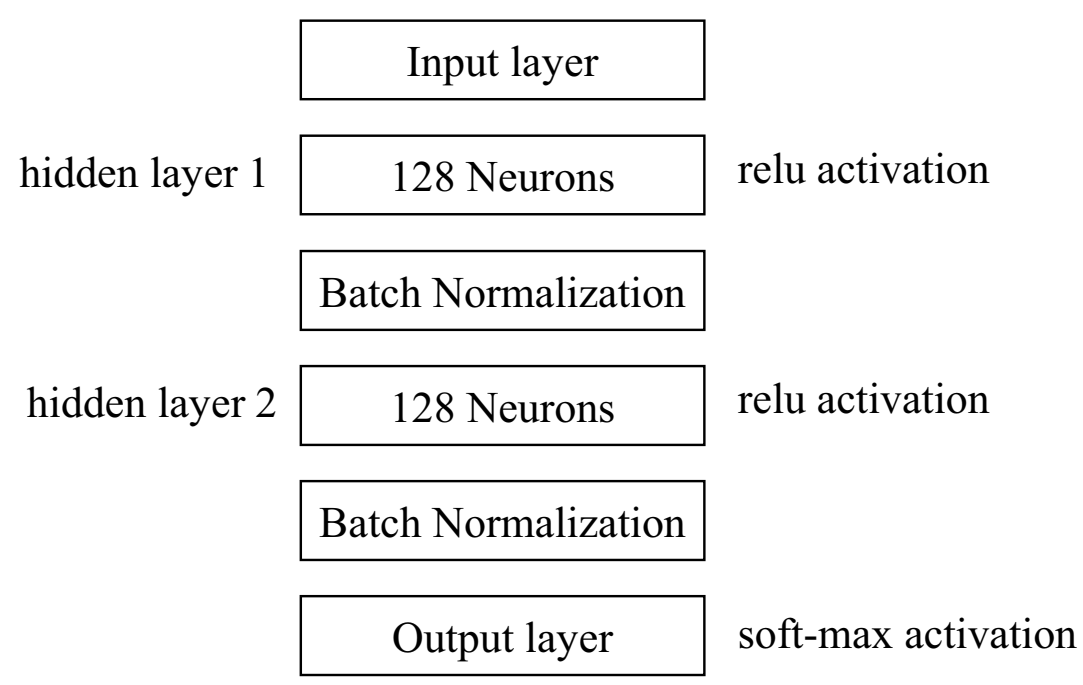

**Fig. 5** Our Specified MLP Classifier

**Table 1** Hardware and environment specification

| Unit | Description |
| --- | --- |
| Processor | AMD Ryzen 7 2700 |
| RAM | 16 GB |
| GPU | AMD RX580 |
| Operating System | Ubuntu 20.04.4 LTS |
| Packages | Tensorflow 2.4.1, Sklearn 1.0.2, Numpy, Pandas and Matplotlib |

$$T_{IG}(n) = C_1 * n = O(n) \tag{11}$$

$$T_{RF}(n) = C_2 * n = O(n) \tag{12}$$

$$T_{RFE}(n) = C_3 * n * (n-1)/2 = O(n^2) \tag{13}$$

$$T_{IGRF-RFE}(n) = C_1 * n + C_2 * n + C_4 * n(n-1)/2 = O(n^2) \tag{14}$$

where n represents the number of features, $C_1, C_2, C_3, C_4$ represent constant numbers

## Experiments and results

### Hardware and environment setting

Our experiments were conducted on a desktop running with Ubuntu 20.04.4 LTS operating system. The hardware used on the desktop consists of 16GB RAM, AMD Ryzen 7 2700 processor, and an AMD RX580 graphics card. Our experimental environment was based on python 3.8 and the MLP model was created on TensorFlow 2.4.1. Scikit-Learn, Numpy, pandas, matplotlib, and other packages provide data processing, feature selection, and visualization functions for our experiments. Specific hardware and environmental information are presented in Table 1.

### The UNSW-NB15 dataset

For intrusion detection systems based on machine learning methods, datasets play a vital role in the effectiveness against unknown attacks, test performance, and generalizability. The IDS datasets are required to contain a sufficient number of different types of attacks and reflect real-world attack scenarios. A well-known IDS dataset is KDDCup99, which has been widely used in many previous studies. KDDCup99 is a dataset created from MIT Lincoln Laboratory's simulated experiments on cyberattacks to help build machine learning classifiers for intrusion detection [41]. NSL-KDD was created as a cleaned version of KDDCup99, removing the duplicate data in KDD99, and rebuilding the training and testing data [42]. However, these two datasets met with criticism that they do not meet the network security requirements of today due to the lack of modern attack types, the imbalanced distribution of training and test sets, and the lack of support for some common network protocols [43].

For the shortcomings of KDDCup/NSL-KDD, Moustafa and Slay [43] created a more complex intrusion detection dataset named UNSW-NB15 to reflect better modern attacks and protocols. UNSW-NB15 is a dataset extracted from 100 GB of normal and modern attack traffic by researchers at the Australian Centre for Cyber Security(ACCS) using the IXIA tool. The complete UNSW-NB15 dataset contains 2.5 million records of data, covering one normal class and nine attack classes which are Analysis, Backdoor, DoS, Exploits, Fuzzers, Generic, Reconnaissance, Shellcode, and Worms. The original data consists of 49 features which can be divided into six groups: flow features, basic features, content features, time features, additional generated features, and labeled features.

The creator of the dataset also provided a 10% partitioned dataset, split into a training set (175,341 records) and a test set (82,332 records) (see Table 2). The statistical distributions of the training and test set samples have been verified to be highly correlated, which means the partitioning is reliable for the machine learning model [44]. There are some minority classes: Analysis, Backdoor, Shellcode, and Worms, whose proportion is less than 2%. In the 10% dataset, a few meaningless features were removed, and the number of features was reduced to 42, including 38 numerical features and 3 categorical features (see Table 3). In this research, we used the 10% dataset for classification.

**Table 2** Records of 10% UNSW-NB15 dataset

| Class | Training dataset | Test dataset |
|---|---|---|
| Normal | 56000 | 37000 |
| Generic | 40000 | 18871 |
| Exploits | 33393 | 11132 |
| Fuzzers | 18184 | 6062 |
| DoS | 12264 | 4089 |
| Reconnaissance | 10491 | 3496 |
| Analysis | 2000 | 667 |
| Backdoor | 1746 | 583 |
| Shellcode | 1133 | 378 |
| Worms | 130 | 44 |
| Total | 175341 | 82332 |

**Table 3** UNSW-NB15 feature data types

| No. | Feature | Dtype | No | Feature | Dtype |
|---|---|---|---|---|---|
| 0 | dur | float64 | 22 | dwin | int64 |
| 1 | proto | object | 23 | tcprtt | float64 |
| 2 | service | object | 24 | synack | float64 |
| 3 | state | object | 25 | ackdat | float64 |
| 4 | spkts | int64 | 26 | smean | int64 |
| 5 | dpkts | int64 | 27 | dmean | int64 |
| 6 | sbytes | int64 | 28 | trans_depth | int64 |
| 7 | dbytes | int64 | 29 | response_body_len | int64 |
| 8 | rate | float64 | 30 | ct_srv_src | int64 |
| 9 | sttl | int64 | 31 | ct_state_ttl | int64 |
| 10 | dttl | int64 | 32 | ct_dst_ltm | int64 |
| 11 | sload | float64 | 33 | ct_src_dport_ltm | int64 |
| 12 | dload | float64 | 34 | ct_dst_sport_ltm | int64 |
| 13 | sloss | int64 | 35 | ct_dst_src_ltm | int64 |
| 14 | dloss | int64 | 36 | is_ftp_login | int64 |
| 15 | sinpkt | float64 | 37 | ct_ftp_cmd | int64 |
| 16 | dinpkt | float64 | 38 | ct_flw_http_mthd | int64 |
| 17 | sjit | float64 | 39 | ct_src_ltm | int64 |
| 18 | djit | float64 | 40 | ct_srv_dst | int64 |
| 19 | swin | int64 | 41 | is_sm_ips_ports | int64 |
| 20 | stcpb | int64 | 42 | attack_cat | object |
| 21 | dtcpb | int64 | 43 | label | int64 |

## Data pre-processing

In this section, we discussed the procedure and methods we use for the data pre-processing process.

### *Cleaning*

In the training and test sets provided by UNSW-NB15, there are 44 original features. 42 of them are meaningful features and 2 features are the class labels of the attack. ‘attack_cat’ is a multi-class label and ‘label’ is a binary-class label. As our MLP model is designed to perform multi-classification for intrusion detection, ‘label’ was removed. In addition, we also cleaned 44 rows with null values in the dataset.

### *Minority removal*

Extremely imbalanced datasets can have a negative impact on machine learning performance. Since the imbalanced dataset is not the focus of this study, we removed 4 minority classes: ‘Analysis’, ‘Backdoor’, ‘Shellcode’, ‘Worms’, which accounted for only 1.141%, 0.996%, 0.646%, and 0.074% of the training set.

### *Oversampling*

We observed that the proportions of data samples for different classes were the same in the given training set and test set while the normal class has the largest difference between them. The normal class accounts for only 32.9% in the training set, while 45.9% instances are the normal class in the test set (see Fig. 6). Dataset imbalance could cause a serious performance issue that affects the training process of MLP models. In UNSW-N15, the proportion of a normal class in the training set is much less than the proportion of a normal class in the test set, which may lead to an overfitting issue towards abnormal classes. To address this issue, we adopted an oversampling technique by double sampling the normal class so that the proportion of the normal class reaches 49.5%.

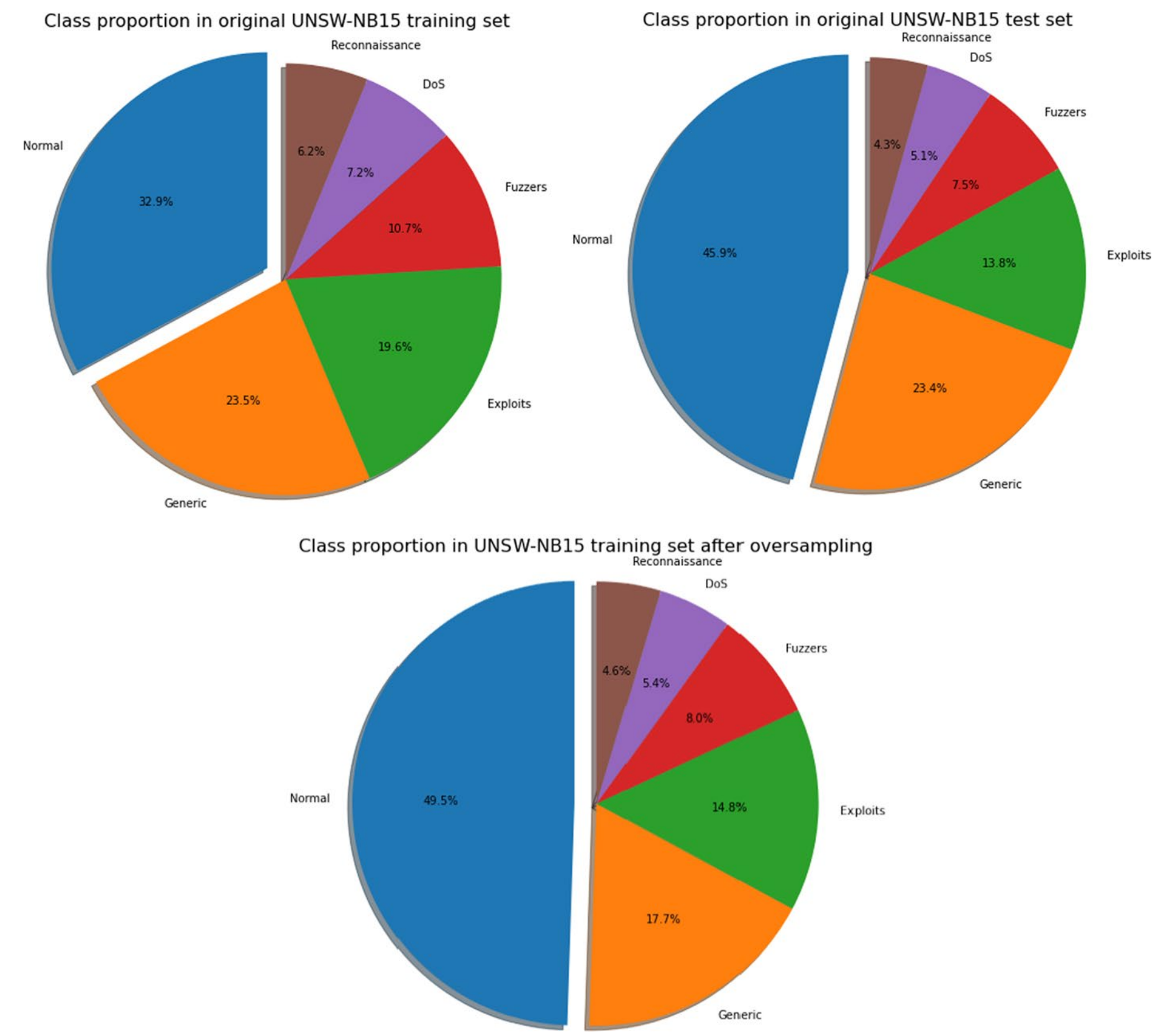

**Fig. 6** Class proportion before and after oversampling

#### *One-hot encoding*

There are three categorical features in the dataset: 'service', 'proto', and 'state', which contain 13, 9133 nominal values respectively. These features were transformed using one-hot encoding, making each nominal value a binary feature.

#### *Normalization*

Normalization can unify the value range of each feature and eliminate the bias during MLP model training caused by different value scales. We used MinMax Normalization to convert the range of feature values between 0 and 1 [45]. As defined in Eq. 15, the new value is calculated by the difference between the min value divided by the scale size.

$$x_i' = \frac{x_i - \min(x_i)}{\max(x_i) - \min(x_i)} \tag{15}$$

where $x_i$ represents the $i$th feature vector, $\min(x_i)$ returns the minimum value of the vector and $\max(x_i)$ returns the maximum value of the vector.

### Training, validation, and test set preparation

In Fig. 7, we applied PCA to the original training and test sets provided by UNSW-NB15, reduced them to three dimensions, and visualized their distribution. The distribution of different classes can be seen in the visualization of PCA latent space, which increases the interpretability of the data. Although PCA visualization cannot represent all dimensions of the data, it can be found that in three-dimensional space, there is a lot of overlap between different types of attack and normal classes. On the other hand, from the 3-dimensional visualization of the training set and the test set, it can be seen that the spatial distribution of the training set and the test set in some areas is not the same.

Machine learning usually divides data sets to use for different purposes. The training set is used to fit the model, the validation set is used to estimate the loss in training, and the test set is used to verify the performance of the model [46]. These three sets suppose to contain separate data samples to avoid biased performance caused by data leakage. The UNSW-NB15 dataset does not provide a separate validation set, as such most previous studies holdout a validation set from the training set. However, in the PCA visualization (see Fig. 7), we observed that the original training set and test set have different distributions in 3-dimensional space, so the model fitted based on the training set may not reflect the performance of the test set. In this case, the model may overfit the special distribution of the training set and cannot generalize well. The validation set usually needs to be the same distribution as the test set to correctly estimate the training loss of the model [47]. As a result, in our study, we split the original test set to construct a new validation set and test set (see Table 4). The new validation set and test set have the same distribution to help the model avoid overfitting. In Table 4, the training, validation, and test set do not overlap, and the ratio of the three datasets is 68:16:16.

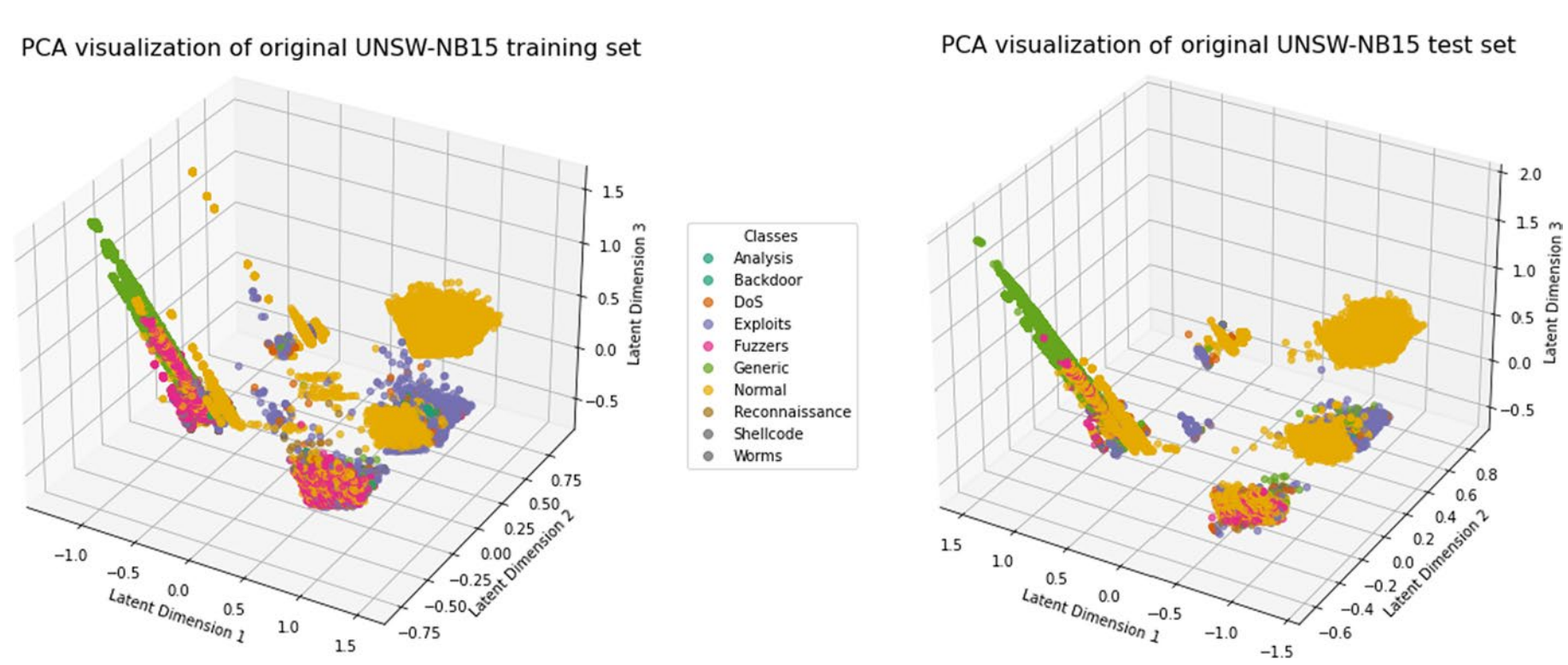

**Fig. 7** The PCA visualization of original training and test set

**Table 4** Records of training, validation and test set to be used in our model

| Class | Training set | Validation set | Test set |
|---|---|---|---|
| Normal | 56000 | 18500 | 18500 |
| Generic | 40000 | 9436 | 9435 |
| Exploits | 33393 | 5566 | 5566 |
| Fuzzers | 18184 | 3031 | 3031 |
| DoS | 12264 | 2044 | 2045 |
| Reconnaissance | 10491 | 1748 | 1748 |
| Total | 170332 | 40325 | 40325 |

**Table 5** Simplified confusion matrix

| Labeled samples | | Predicted cass | |
|---|---|---|---|
| | | Positive | Negative |
| Actual Class | Positive | TP | FN |
| | Negative | FP | TN |

#### Evaluation metrics

As our work is a multi-classification task, we used accuracy, recall, precision, false positive rate (FPR), f1 score, and AUC-ROC curve as our performance metrics. Table 5 presents a simplified confusion matrix that differentiates the classification results. Based on the one versus all principle, there are generally four cases in machine learning classification tasks, where:

- True Positive (TP): represents correctly classified positive samples
- False Negative (FN): represents incorrectly classified positive samples
- False Positive (FP): represents incorrectly classified negative samples
- True Negative (TN): represents correctly classified negative samples

Accuracy as defined in Eq. 16 calculates the ratio of correctly classified samples to all samples.

$$Accuracy = \frac{TP + TN}{TP + TN + FP + FN} \tag{16}$$

Recall as defined in Eq. 17 calculates the ratio of correctly classified positive samples to all samples that were supposed to be positive.

$$Recall(TruePositiveRate) = \frac{TP}{TP + FN} \tag{17}$$

Precision, as defined in Eq. 18, calculates the ratio of actually classified positive samples to all samples that are predicted to be positive.

$$Precision = \frac{TP}{TP + FP} \tag{18}$$

False positive rate (FPR) as defined in Eq. 19 calculates the ratio of incorrectly classified positive samples to all samples that were supposed to be negative.

$$FPR = \frac{FP}{TN + FP} \tag{19}$$

F1 score as defined in Eq. 20 calculates the harmony mean of recall and precision. It can be used as a performance metric to solve the defects of recall and precision when faced with multi-class imbalanced data.

$$F1 = 2 \times \left( \frac{Precision \times Recall}{Precision + Recall} \right) \tag{20}$$

The Receiver operating characteristic (ROC) curve shows the FPR and TPR of the model prediction at different thresholds. The area under the ROC curve (AUC) as defined in Eq. 21 calculates the area under the ROC, and it can be used to judge the performance of the model.

$$AUC_{ROC} = \int_{0}^{1} \frac{TP}{TP + FN} d \frac{FP}{TN + FP} \tag{21}$$

## Results

Before the ensemble feature selection with IG and RF, we removed the duplicate samples in the training set to avoid overfitting features. Then, we applied the information gain and a random forest classifier with 1000 trees on the training set to obtain the importance ranking of 39 numeric features (see Figs. 8 and 9). There are some low importance features in IG ranking and RF importance ranking, which may degrade the performance of the model. We choose 0.25 and 0.02 as the thresholds for two feature selection methods respectively to filter important features. Therefore, in IG ranking, features with an importance score greater than 0.25 were retained while features with an importance score greater than 0.02 were retained in RF importance ranking. After removing unimportant features from these two metrics respectively, two feature subsets were obtained.

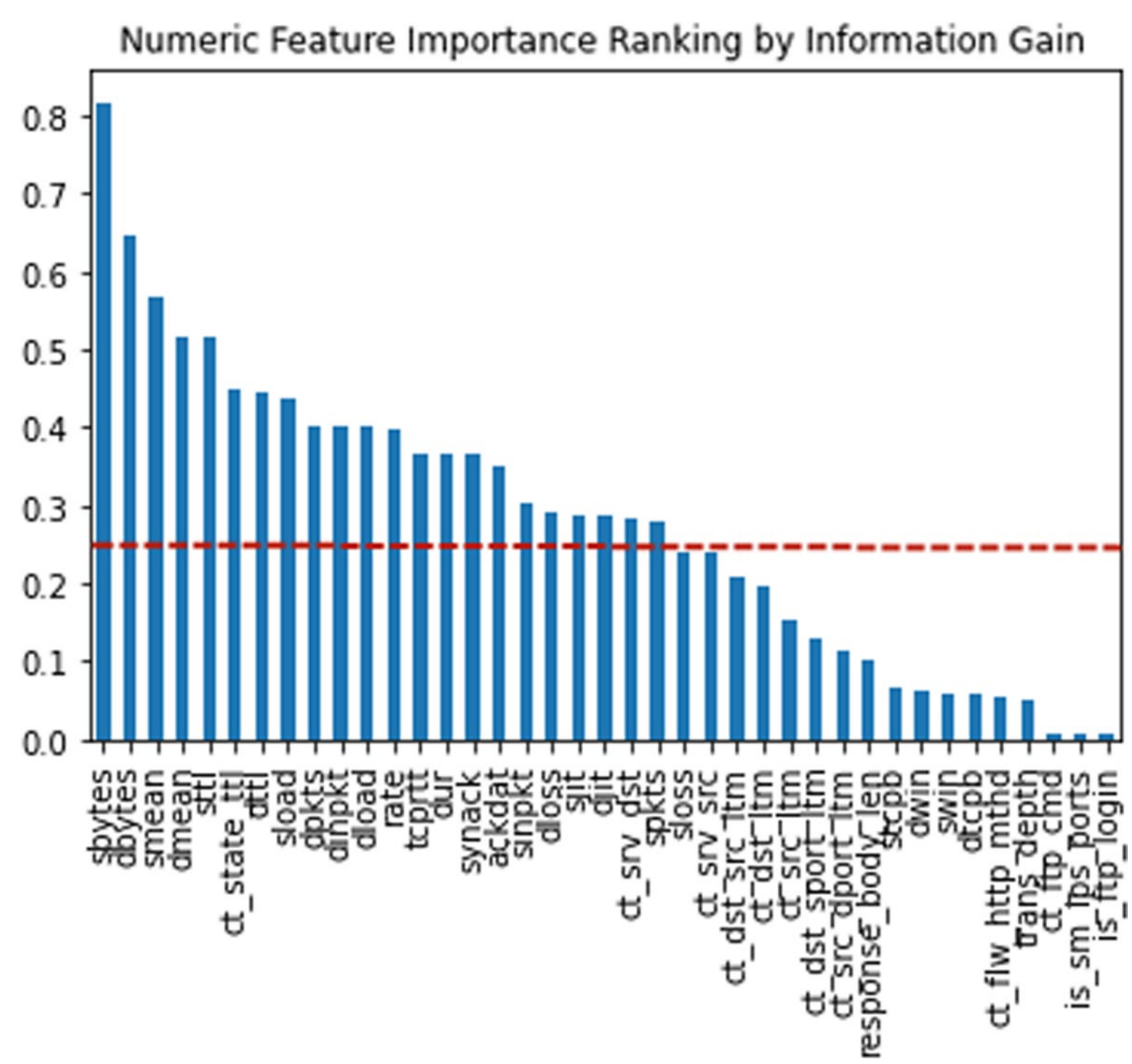

**Fig. 8** Numeric feature importance ranking by information gain

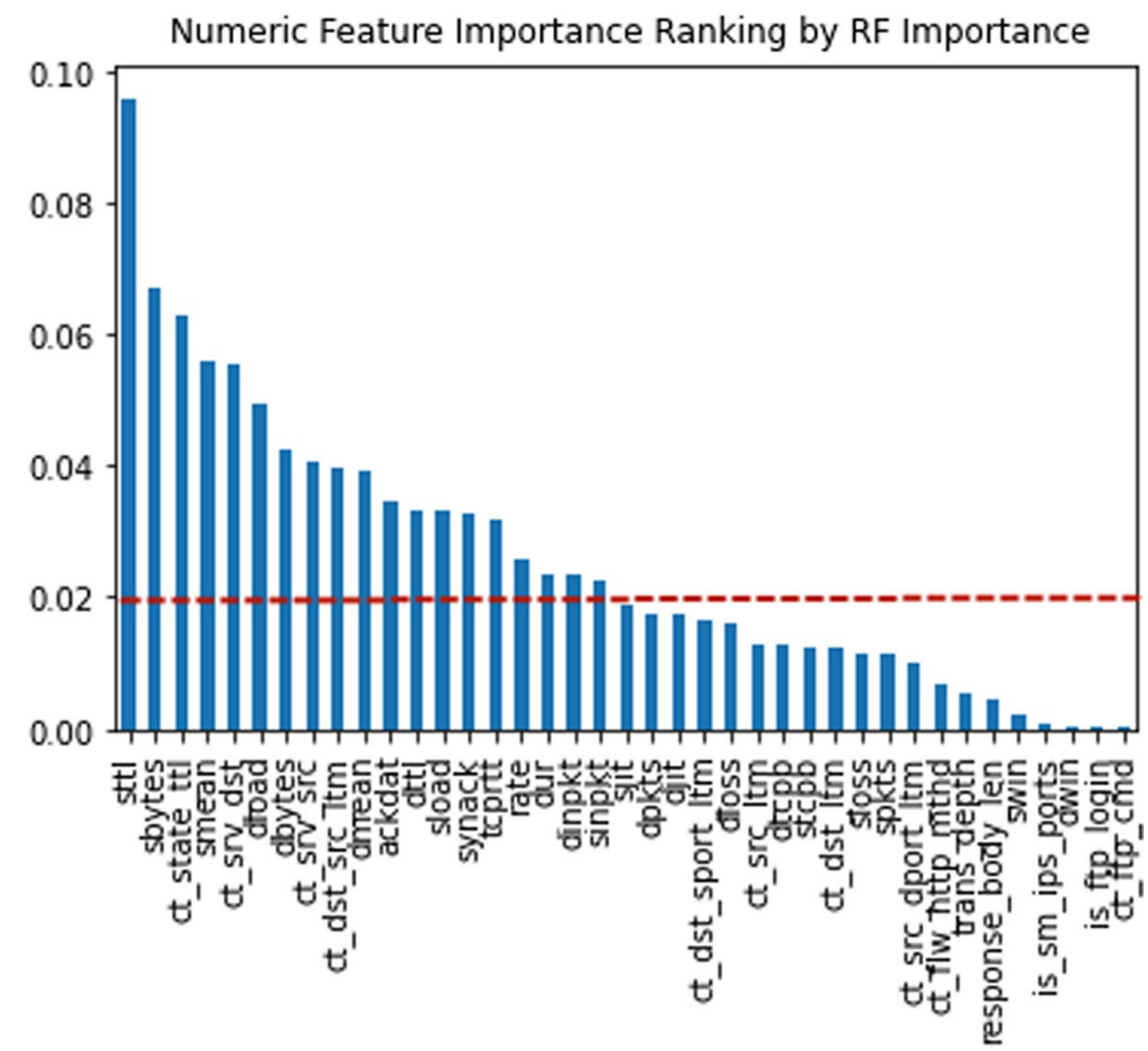

**Fig. 9** Numeric feature importance ranking by random forest

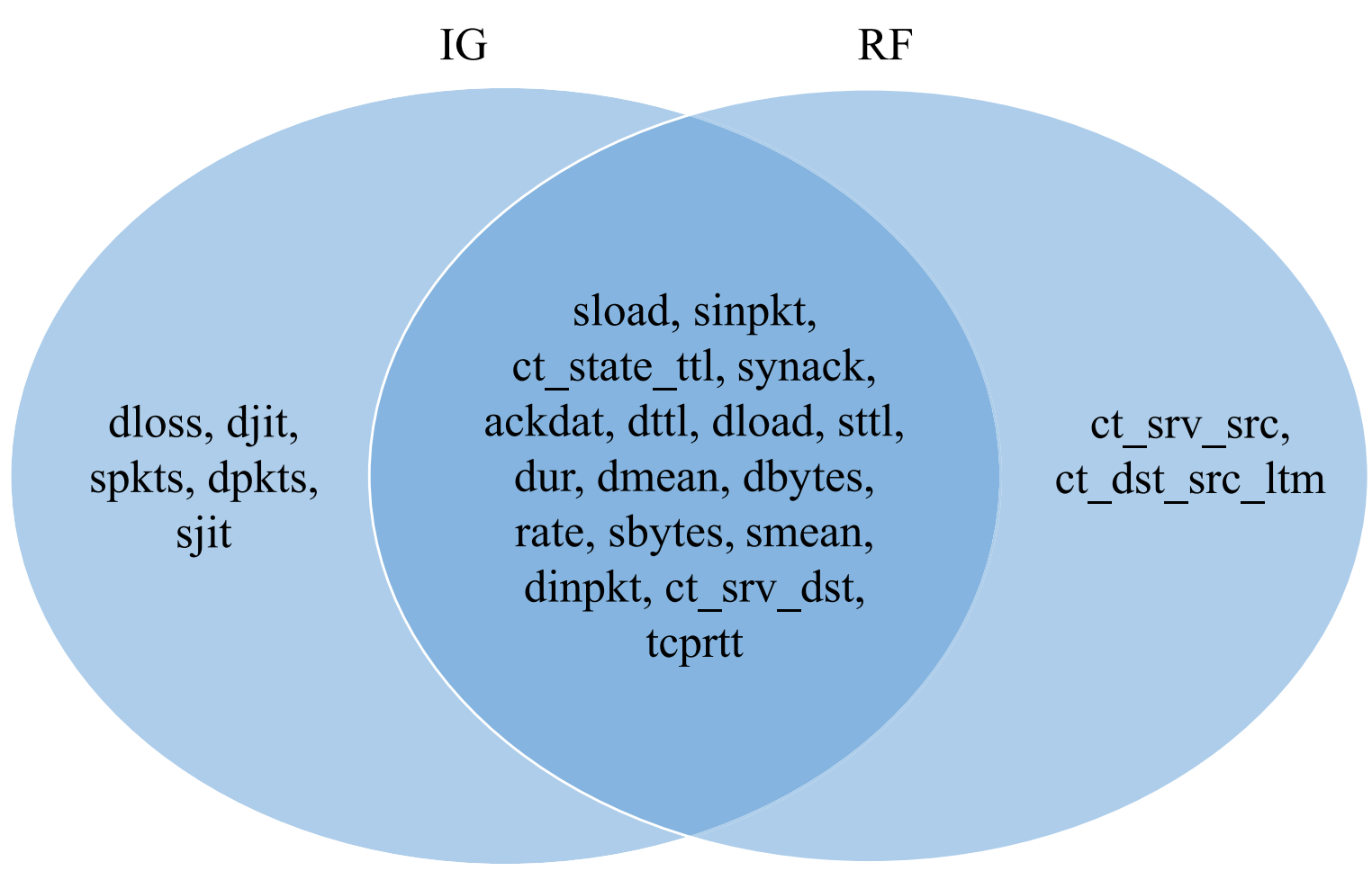

**Fig. 10** Numeric feature union set of IG and RF feature selection

22 features were retained by IG feature selection and 19 features were retrained by RF importance feature selection. To take their union set in Fig. 10, it can be seen that they have 17 features in common and their union set have 24 numeric features which can be used as a part of the reduced feature subset for further feature reduction in the second step.

We applied the 24 numerical features obtained in the IGRF ensemble step and 3 categorical features to the wrapper-based RFE feature selection method. 10 random seeds from 2022 to 2031 were chosen to average the score in the evaluate_elimination function for each elimination. In the experiment, the patient parameter in our proposed model was set to 5, which means that the RFE process is stopped if the performance does not improve in the cumulative five iterations. After applying our hybrid feature selection method on UNSW-NB15, 23 important features were finally selected including 20 numerical features and 3 categorical features (see Table 6). In Fig. 11, the confusion matrix of multi-classification is displayed, where the horizontal axis is the predicted label and the vertical axis is the true label. It can be seen that there are some misclassifications among different classes. DoS, Fuzzer, and Reconnaissance were often misclassified as Exploits. Approximately 88.31% of DoS samples, 27.78% of Fuzzer samples, and 18.82% of Reconnaissance samples were misclassified as Exploits, which might have been the reason for their poor performances. In addition, 889 samples of the Fuzzer class were misclassified as normal while 966 normal samples were misclassified as Fuzzer class.

In Table 7, the performance of our IDS model is presented. In multi-classification, the MLP model based on the IGRF-RFE feature selection method has an accuracy of 84.24%. Since the UNSW-NB15 dataset has multiple imbalanced classes, the f1 score is a better measure to use for the performance of each class. The Generic attack has the highest f1 score of 98.20%, followed by the normal class with 93.11%. DoS and Fuzzer attacks have lower f1 scores of 11.09% and 42.26% respectively, which may be the insufficient samples in the UNSW-NB15 dataset. DoS class only occupies 5.4% of the training set and Fuzzers only accounts for 8.0%, which may make the MLP model can not fit them well when training. Although the Exploits and Reconnaissance classes do not have as many samples as Generic and Normal, they have f1 scores of 72.55% and 78.83%. Moreover, a lower false positive rate (FPR) also reflects one aspect of the model performance. Our model has a weighted FPR of 0.0403, which means that only about 4% of negative samples are misclassified to be positive samples. The Generic attack class has the lowest FPR of 0.0005, while the Exploits has the worst FPR of 0.993. Also, although DoS has the lowest f1 score, it has the second-lowest FPR of 0.0062.

**Table 6** Selected features by IGRF-RFE

| No. | Feature | Dtype | No. | Feature | Dtype |
|---|---|---|---|---|---|
| 0 | dur | float64 | 14 | dloss | float64 |
| 1 | proto | object | 15 | sinpkt | float64 |
| 2 | service | object | 16 | dinpkt | float64 |
| 3 | state | object | 18 | djit | float64 |
| 4 | spkts | float64 | 23 | tcprtt | float64 |
| 5 | dpkts | float64 | 24 | synack | float64 |
| 6 | sbytes | float64 | 25 | ackdat | float64 |
| 7 | dbytes | float64 | 26 | smean | float64 |
| 8 | rate | float64 | 27 | dmean | float64 |
| 9 | sttl | float64 | 31 | ct_state_ttl | float64 |
| 10 | dttl | float64 | 35 | ct_dst_src_ltm | float64 |
| 12 | dload | float64 | | | |

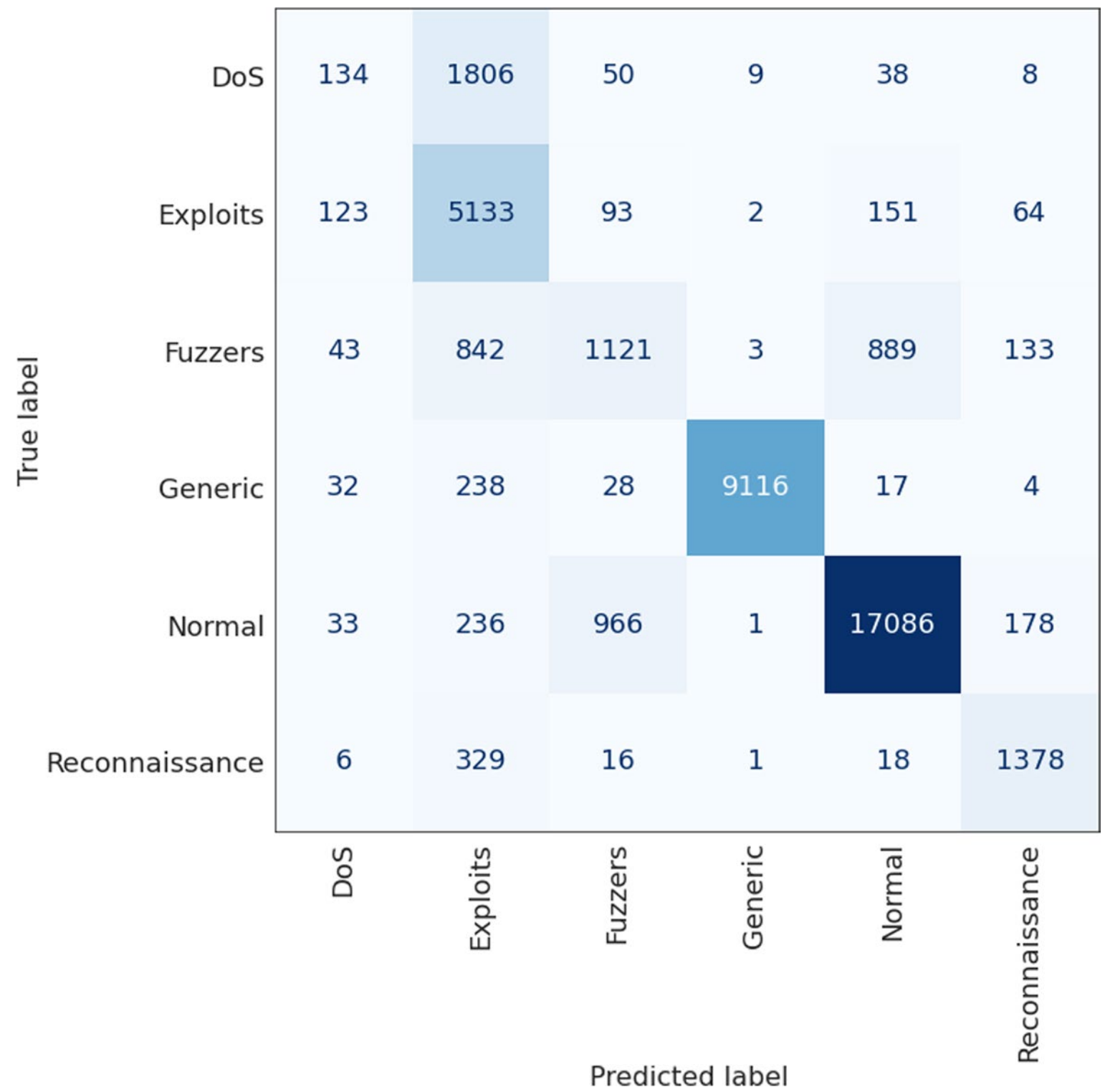

**Fig. 11** Confusion matrix of our results

**Table 7** Evaluation metrics of multi-classification of UNSW-NB15

| | Precision | Recall | F1 Score | FPR | Accuracy |
|---|---|---|---|---|---|
| DoS | 0.3612 | 0.0655 | 0.1109 | 0.0062 | 84.24% |
| Expl. | 0.5980 | 0.9222 | 0.7255 | 0.0993 | |
| Fuzz. | 0.4930 | 0.3698 | 0.4226 | 0.0309 | |
| Gene. | 0.9982 | 0.9662 | 0.9820 | 0.0005 | |
| Norm. | 0.9388 | 0.9236 | 0.9311 | 0.0510 | |
| Recon. | 0.7807 | 0.7883 | 0.7845 | 0.0100 | |
| Weighted Avg. | 0.8360 | 0.8424 | 0.8285 | 0.0403 | |

In Fig. 12, we applied the one vs all methodology to generate theReceiver Operating Characteristic (ROC) curve for each class, which can help understand the quality of the predicted probability. Generic and normal classes have the higher AUCs (area under the ROC curve) of 1 and 0.99 respectively. However, DoS and Fuzzer classes have the lower AUCs of only 0.95 and 0.89. Overall, the multi-class ROC curve reflects a good performance of our model's detection capability.

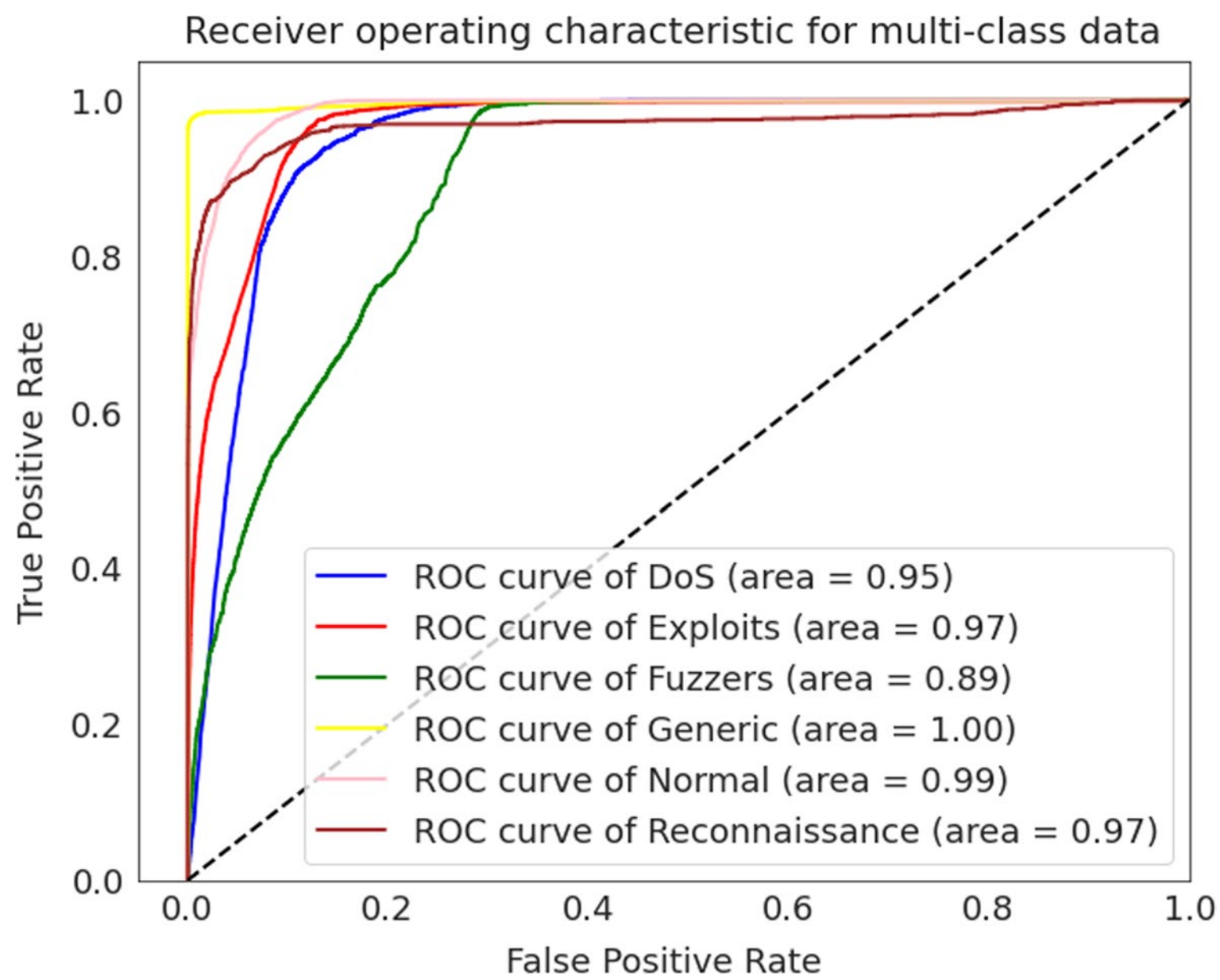

**Fig. 12** ROC (receiver operating characteristic) curve of multi-classification

**Table 8** Performance of different feature subsets

| Subsets | Num. | Precision (%) | Recall (%) | F1 score (%) | Accuracy (%) |
|---|---|---|---|---|---|
| All Features | 42 | 80.37 | 82.25 | 80.22 | 82.25 |
| IG | 25 | 82.24 | 83.13 | 81.49 | 83.13 |
| RF | 22 | 82.43 | 83.42 | 81.62 | 83.42 |
| IG &RF Uni. | 27 | 83.30 | 80.60 | 80.36 | 80.60 |
| IG &RF Inter. | 20 | 81.84 | 82.90 | 81.67 | 82.90 |
| IGRF-RFE | 23 | 83.60 | 84.24 | 82.85 | 84.24 |

## Comparison

From Table 8, we compared our results with the performance of different feature selection methods and feature subsets used in our experiments. It can be seen that the feature subset selected by our IGRF-RFE method performs better than other feature subsets in the Table. Our feature selection method improves nearly 2% in accuracy and 2.6% in weighted f1 score over using all features. Furthermore, we evaluate the performance of feature subsets using standalone IG and RF and their union and intersection sets. Using IG and RF's feature subsets as well as their intersection improves the performance of the model, but they are still lower than our methods in every evaluation metric.

We also compared the performance of our proposed method with other similar previous works (see Table 9). Among similar works using MLP models, our MLP model with the IGRF-RFE feature selection method achieves better performance in both f1 score and accuracy. Our hybrid feature selection method obtains 23 important features and outperforms the standalone IG method or some other tree-based feature selection methods. It is worth mentioning that although our performance is lower than the 95.2% accuracy achieved by Prasad et al's work [27], it is because different studies use varying amounts of data for UNSW-NB15. Our study used the 10% pre-partitioned dataset from UNSW-NB15's author which is validated by statistical distributions, and our results are still competitive among similar methods.

**Table 9** Comparison results of other methods on UNSW-NB15 dataset

| Work | Classifier | FS Method | No. of features | F1 score(%) | Accuracy(%) |
|---|---|---|---|---|---|
| Kasongo and Sun [25] | ANN | XGBoost | 19 | 77.28 | 77.51 |
| Roy and Singh [48] | MLP | IG | 20 | - | 84.1 |
| Kasongo and Sun [49] | FFDNN | WFEU | 22 | - | 77.16 |
| Eunice et al. [26] | DNN | DT | 20 | - | 82.1 |
| Moustafa and Slay [44] | ANN | - | 42 | - | 81.34 |
| Proposed method | MLP | IGRF-RFE | 23 | 82.85 | 84.24 |

## Conclusion and future work

In this paper, we proposed a hybrid feature selection method IGRF-RFE for MLP-based intrusion detection systems and applied it to a modern IDS dataset UNSW-NB15. IGRF-RFE consists of two feature reduction steps including IGRF ensemble feature selection and recursive feature elimination with MLP. In IGRF ensemble feature reduction, 24 important numerical features were obtained according to the importance ranking of numerical features ranked by Information Gain IG) and Random Forest (RF) methods. After feeding 24 numerical features and 3 categorical features into the wrapper-based RFE algorithm, we obtained an optimal feature subset with 20 numerical features and 3 categorical features for our MLP model. Our hybrid feature selection approach has a worst-case computational complexity of $O(n^2)$, which is equivalent to that of the normal RFE algorithm. By introducing patient parameter $p$, our algorithm can stop earlier to save computational resources. The experimental results showed that our feature selection method could achieve an accuracy of 84.24% and a weighted f1 score of 82.85%, which was better than standalone IG and RF feature selection methods as well as other similar previous work.

The results also show that using the proposed IGRF-RFE feature selection method can effectively select important features and improve the performance of intrusion detection systems. At the same time, the method can also be applied to the feature selection of other structured datasets. In the future, we plan to apply our proposed feature selection method to different intrusion detection datasets with advanced re-sampling techniques as well as with other machine learning models [50].

**Acknowledgements**
Not applicable.

**Author contributions**
Conceptualization, YY and JJ-J; methodology, YY and JJ-J; software, YY; formal analysis, YY; writing—original draft preparation, YY and JJ-J; writing—review and editing, YY, JJ-J, WX, AS, JZ, FS and JK; funding acquisition, JJ-J; project administration, JJ-J. All authors have read and agreed to the published version of the manuscript. All authors read and approved the final manuscript.

**Funding**
This work is supported by the Cyber Security Research Programme-Artificial Intelligence for Automating Response to Threats from the Ministry of Business, Innovation, and Employment (MBIE) of New Zealand as a part of the Catalyst Strategy Funds under the Grant Number MAUX1912.

**Availability of data and materials**
The dataset used in this research is available on https://www.unsw.adfa.edu.au/unsw-canberra-cyber/cybersecurity/ADFA-NB15-Datasets/.

## Declarations

**Ethics approval and consent to participate**
Not applicable.

**Consent for publication**
The authors agree to publish this paper.

**Competing interests**
The authors declare that they have no competing interests.

## Publisher's Note
Springer Nature remains neutral with regard to jurisdictional claims in published maps and institutional affiliations.